\pdfoutput=1
\documentclass[conference]{IEEEtran}
\IEEEoverridecommandlockouts
\usepackage{graphicx}
\usepackage[numbers]{natbib}
\usepackage{multicol}
\usepackage{bm}
\usepackage[bookmarks=true]{hyperref}
\usepackage[acronym,toc]{glossaries}

\usepackage[font=small]{caption}

\newacronym{CBF}{CBF}{Control Barrier Function}
\newacronym{NCBF}{NCBF}{Neural Control Barrier Function}

\usepackage{hyperref}
\usepackage{url}
\usepackage{graphicx}
\usepackage{float}

\usepackage{algorithm}
\usepackage{amsmath}
\usepackage{amssymb}
\usepackage[noend]{algpseudocode}
\usepackage[T1]{fontenc}
\usepackage[utf8]{inputenc}

\usepackage{booktabs}
\usepackage{tabularx}
\usepackage{array}
\usepackage{xcolor}
\usepackage{dsfont}

\usepackage{subfigure}

\title{Safe Reinforcement Learning with Receding-Horizon Safety Critic and Differentiable Action Shields}

\author{
  Siwei Ju$^{1, *}$,
  Jan Peters$^{1,2,3,4}$,
  and Oleg Arenz$^{1}$%
\thanks{
\newline $^{1}$Department of Computer Science, Technical University of Darmstadt, Germany. 
\newline $^{2}$Hessian.AI.
\newline $^{3}$German Research Center for AI (DFKI), Research Department: Systems AI for Robot Learning.
\newline $^{4}$Robotics Institute Germany (RIG).
\newline $^{*}$ Corresponding author: {\tt\small siwei@robot-learning.de} 
}
}

\begin{document}

\maketitle

\begin{abstract}
Despite the Markov assumption, safety failures in robot reinforcement learning rarely occur instantaneously; they typically emerge over several steps as the policy enters unfamiliar states under disturbances and domain shift. We address this limitation by learning a \emph{receding-horizon safety critic} that predicts whether a state will remain non-terminating over the next receding horizon, enabling anticipatory detection of impending locomotion failures. The critic is trained from rollout-derived horizon labels with a composite objective that combines horizon classification, control-barrier-function-inspired temporal consistency, and smoothness regularization, and we aggregate an ensemble of critics with a conservative Conditional Value at Risk estimate to account for uncertainty. Building on the learned critic, we derive a closed-form, differentiable CBF-style safety filter that minimally modifies the policy action to improve predicted safety, and integrate it into PPO with an anticipation loss to reduce policy–filter mismatch during training. In simulated quadruped locomotion, our method improves robustness under action noise, external perturbations, and terrain shifts, reducing catastrophic failures while maintaining competitive returns.
\end{abstract}

\section{Introduction}

Reinforcement learning (RL) has enabled increasingly agile legged locomotion, yet safety remains a central obstacle to deploying learned controllers beyond narrow training conditions. In practice, many locomotion failures are delayed: a destabilizing contact, a slight foot scuff, or a suboptimal recovery step may not cause an immediate violation, but can drive the robot into an unrecoverable state several steps later. This makes safety inherently horizon-dependent in locomotion, especially under disturbances and terrain-induced contact variability.

A common approach to safety in continuous control is to enforce state constraints using control barrier functions (CBFs)~\cite{ames_control_2017} and differentiable safety layers that project actions through constrained optimization~\cite{agrawal_differentiable_2019, xiao_barriernet_2021}. Data-driven CBF variants often learn uncertainty or residual terms to compensate modeling error~\cite{taylor_learning_2019, emam_data-driven_2021}, while neural CBFs (NCBFs) parameterize the certificate itself to reduce manual design effort~\cite{zhang_exact_2023, so_how_2023, robey_learning_2020, tabbara_learning_2025, yang_learning_2025, long_learning_2021, agrawal_neural_2025}. In safe RL, CBF-style shields are frequently combined with actor--critic training~\cite{emam_safe_2022, cheng_end--end_2019, du_reinforcement_2023, xie_certificated_2025, yang_cbf-rl_2025} or treated as part of the environment dynamics~\cite{safety_filter_rl_2025}. Despite their success, these methods face persistent challenges in locomotion: (i) computational overhead from per-step optimization and the associated training-time scalability limits, (ii) \emph{myopia} from one-step constraints on the safety layer that may not anticipate multi-step failures, and (iii) \emph{policy--filter mismatch}, where the executed (filtered) actions induce a distribution shift that is not always accounted for during policy optimization.

Predictive safety filters address myopia by enforcing safety over a horizon, typically relying on known or nominal dynamics models and uncertainty representations~\cite{wabersich_predictive_2021, wabersich_predictive_2023, wabersich_data-driven_2023}. In parallel, safe RL has explored learning feasibility or long-horizon safety signals that resemble critics/value functions~\cite{long_term_safety_puze2024, value_function_cbf} and structured safety enforcement that avoids generic QPs when dynamics are known~\cite{liu_atacom_2024}. For robot control, complementary approaches predict failure and switch to recovery behaviors (e.g., safe-fall policies)~\cite{meng_safefall_2025, agile_but_safe_2024}, or deploy CBF shields on complex platforms such as humanoids~\cite{yang_shield_2025, yang_cbf-rl_2025}. These strands collectively indicate the need for safety mechanisms that are horizon-aware, computationally lightweight, and compatible with end-to-end policy learning under distribution shift.

We propose a \emph{receding-horizon safety critic} that directly predicts whether the system will remain non-terminated over the next $H$ steps from a state. The critic is trained from rollouts using horizon-wise binary labels, shaped by a temporal consistency prior inspired by discrete-time CBF conditions, and regularized for smoothness to improve robustness under distribution shift. To make the safety signal actionable for control, we derive a differentiable safety layer that minimally corrects the policy action via a \emph{single-constraint} projection with a closed-form solution, and we use a slack variable scaled by the ensemble uncertainty. Because the safety layer is differentiable, we integrate it into PPO and introduce a safety anticipation loss so the policy learns to reduce interventions and align with the horizon-safety objective.

We evaluate the approach on quadruped locomotion under controlled domain shifts, including stochastic action noise and external perturbations on both flat and heightfield terrains. Across disturbance intensities, the proposed method substantially improves success rates while maintaining competitive returns compared to PPO with domain randomization and PPO augmented with an NCBF-style safety layer. An ablation on the horizon length reveals a clear trade-off: longer horizons better mitigate myopic failures but are harder to train and can become over-conservative under shift.

We summarize our contribution as follows:
\begin{itemize}
    \item Receding-horizon safety critic.
    We learn a safety critic that predicts whether the system remains safe over the next $H$ steps from horizon-labeled rollouts.

    \item Structured horizon-safety training.
    We propose a training objective combining horizon classification, temporal consistency, and smoothness regularization with balanced sampling of rare failures.

    \item Closed-form differentiable safety shield.
    We derive a one-constraint projection that minimally corrects the policy action to prevent decreases in predicted safety, enabling end-to-end optimization.

    \item Robust safe locomotion under shift.
    Integrated with PPO via a safety-anticipation loss, our method reduces catastrophic failures and improves robustness under domain shift, outperforming PPO baselines and an NCBF safety layer without hand-designed CBFs.
\end{itemize}

\section{Related Work}
This work intersects barrier-function-based safety, predictive safety filtering, and safe reinforcement learning for legged robots. We review these areas and highlight the gap our approach targets: horizon-aware safety enforcement that is differentiable, model-free at deployment, and compatible with end-to-end policy optimization under distribution shift.

\subsection{Control Barrier Functions, Predictive Safety Layers, and Neural CBFs}
\gls{CBF}s enforce forward invariance of a safe set by constraining control inputs and are a standard tool for safety-critical control~\cite{ames_control_2017}. To handle uncertainty and disturbances, data-driven CBF methods often assume a nominal model and learn learn the residual and uncertainty from data, e.g., reducing model error in the CBF derivative~\cite{taylor_learning_2019} or enabling long-term safe operation under unknown disturbances~\cite{emam_data-driven_2021}.

For neural policies and end-to-end training, CBF constraints are frequently implemented as differentiable optimization layers (typically QPs) so gradients can pass through the safety filter~\cite{ames_control_2017, agrawal_differentiable_2019, xiao_barriernet_2021}. In safe RL, such filters are sometimes treated as part of the environment dynamics~\cite{safety_filter_rl_2025}; examples include CBF layers added to SAC~\cite{emam_safe_2022}, model-based filtering with learned uncertainty (e.g., GP) combined with RL~\cite{cheng_end--end_2019}, and methods that integrate CBF/CLF constraints into actor--critic updates or reward design~\cite{du_reinforcement_2023, xie_certificated_2025, yang_cbf-rl_2025}. A recurring issue is \emph{policy--filter mismatch}: the executed (filtered) action can deviate from the policy output, yet many approaches do not explicitly optimize under the induced distribution shift.

Beyond one-step constraints, predictive safety filters enforce safety over a future horizon, typically requiring nominal or learned dynamics and optional uncertainty models~\cite{wabersich_predictive_2021, wabersich_predictive_2023, wabersich_data-driven_2023}. Finally, \gls{NCBF}s parameterize certificates with neural networks to avoid manual barrier design; recent work studies verification and training of NCBFs~\cite{zhang_exact_2023, so_how_2023}, learning certificates from demonstrations~\cite{robey_learning_2020, tabbara_learning_2025, yang_learning_2025, long_learning_2021}, incorporating physics/PDE structure~\cite{agrawal_neural_2025}, and applications under input limits or from sensory predictors~\cite{liu_safe_2022, gupta_t-cbf_2025}. These tools are increasingly used in robotics, including legged locomotion and manipulation~\cite{arena_stability_2024, grandia_multi-layered_2021, rodriguez_neural_2022, yang_cbf-rl_2025, mcilvanna_reinforcement_2022}, motivating safety mechanisms that remain effective under domain shifts.

\subsection{Relationship to Safe Reinforcement Learning}
Safe RL methods can be broadly grouped into approaches that (i) learn a safety/feasibility signal analogous to a value function, and (ii) enforce constraints via projection or structured action correction.

The former methods learns a scalar signal predictive of future constraint satisfaction to guide exploration or policy improvement. \cite{long_term_safety_puze2024} explicitly targets long-horizon safety under uncertainty, emphasizing that safety is a temporally extended property rather than a purely local certificate. In parallel, \cite{value_function_cbf} connects RL value functions to the barrier-function viewpoint, bridging critics and formal safety reasoning. These perspectives motivate learning safety critics that encode \emph{horizon-aware} feasibility.

Another line avoids per-step QPs by exploiting constraint structure. ATACOM provides a closed-form mechanism to enforce multiple constraints for known dynamics by operating on the constraint manifold, reducing computational overhead relative to generic QP-based filters~\cite{liu_atacom_2024}. Such methods are efficient when modeling assumptions hold, but do not directly address end-to-end integration with policy learning under domain shift, which is central in robotics.

\subsection{Safe Reinforcement Learning for Robot Control}
In robot control, safe RL commonly follows two directions.

(i) Risk prediction with recovery policies.
These approaches train predictors for failure events and switch to a recovery behavior when risk is high. SafeFall trains a falling predictor paired with a safe-fall policy to mitigate catastrophic failures at deployment~\cite{meng_safefall_2025}. Related work emphasizes high-performance locomotion with safety-aware objectives (e.g., collision-free high-speed behaviors)~\cite{agile_but_safe_2024}. While effective, recovery pipelines introduce additional controllers and switching logic, and typically do not provide a single differentiable safety mechanism jointly optimized with the primary policy.

(ii) Safety layers / shields based on barrier functions.
A second family uses CBF-based filtering to minimally modify policy actions to satisfy constraints. Robot-focused work demonstrates CBF-based shields on complex platforms such as humanoids~\cite{yang_shield_2025} and shows that incorporating CBF structure can also benefit learning via reward shaping or auxiliary objectives~\cite{yang_cbf-rl_2025}. Nevertheless, three practical limitations remain: (1) \emph{training--deployment mismatch} when applying the filter only at deployment, it may result in distribution shift; instead using safety filtering also during training can be important~\cite{safe_training_safety_2025, yang_cbf-rl_2025}; (2) \emph{computational overhead} from solving optimization problems online, which hinders real-time deployment and large-scale RL training~\cite{yang_shield_2025}; and (3) local (one-step) constraints and delayed failures: standard CBF/NCBF conditions constrain the instantaneous change of a barrier value, which can be too local in domains where failures emerge over multiple steps (e.g., loss of balance after a disturbance). Predictive safety filters address this by enforcing constraints over a horizon, but they typically lead to more involved multi-step optimization and stronger modeling assumptions.

We address these issues by learning a receding-horizon safety critic and using it as a differentiable safety layer that does not require an explicit dynamics model. This enables end-to-end safety-shielded training to reduce distribution mismatch, avoids per-step QP solves through learned differentiable action correction, and mitigates myopia by optimizing multi-step safety through the horizon-aware critic.

\section{Method}
\newcommand{\E}{\mathbb{E}}
\newcommand{\R}{\mathbb{R}}
\newcommand{\1}{\mathbf{1}}
\newcommand{\grad}{\nabla}
\newcommand{\KL}{\mathrm{KL}}
\newcommand{\Var}{\mathrm{Var}}
\newcommand{\T}{\mathcal{T}}
\newcommand{\V}{\mathcal{V}}
\newcommand{\A}{\mathcal{A}}
\newcommand{\Sset}{\mathcal{S}}
\newcommand{\N}{\mathcal{N}}
\newcommand{\softrelu}{\mathrm{ReLU}}
\newcommand{\softs}{\lambda}
\newcommand{\etaCBF}{\eta}
\newcommand{\hphi}{h_{\phi}}
\newcommand{\piold}{\pi_{\theta_{\text{old}}}}
\newcommand{\pit}{\pi_\theta}
\newcommand{\pibar}{\tilde{\pi}}
\newcommand{\cL}{\mathcal{L}}
\newcommand{\cJ}{\mathcal{J}}
\newcommand{\cD}{\mathcal{D}}

\begin{figure*}[t]
\centering
    \includegraphics[width=\textwidth]{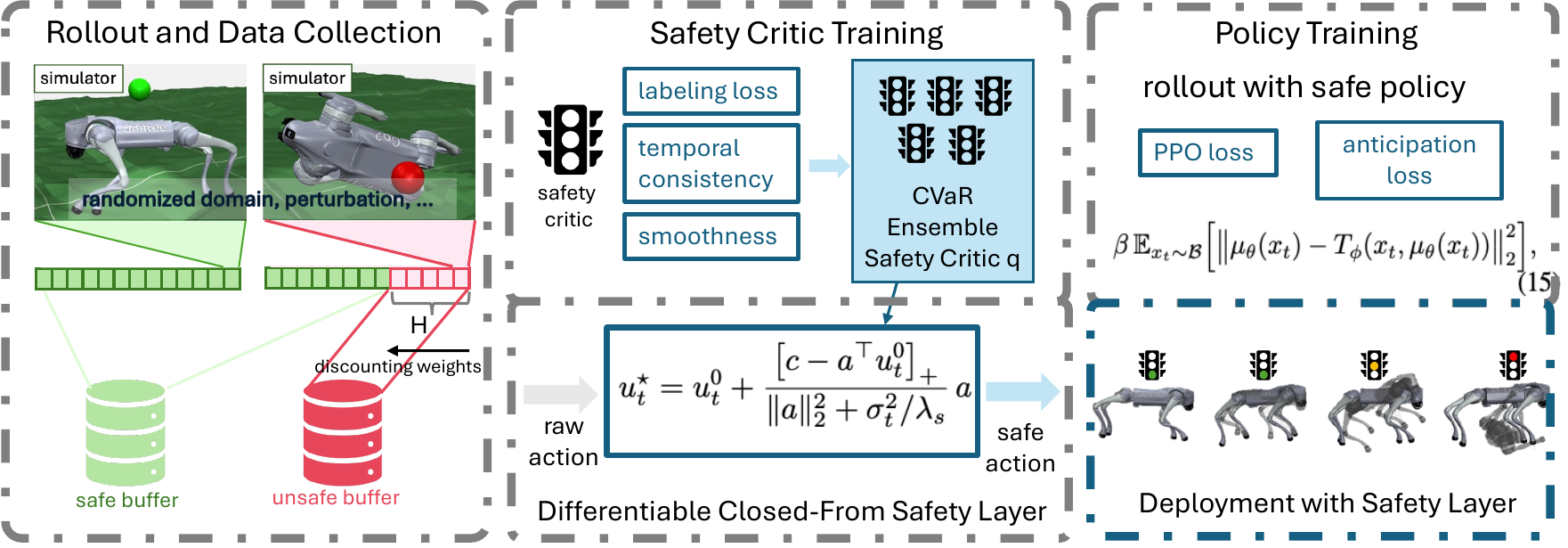}
    \caption{Overview of the training pipeline. Rollouts are collected in simulation under domain randomization and disturbances, and states are labeled by horizon-$H$ safety to populate separate safe/unsafe buffers (with discounted weighting for delayed violations). A predictive safety critic is trained with a composite objective (labeling, temporal consistency, and smoothness), and an ensemble is aggregated via CVaR to obtain a conservative safety estimate and uncertainty $\sigma_t$. At each step, a differentiable closed-form safety layer projects the raw policy action to a minimally modified safe action, which is used both during PPO training (with an anticipation loss to reduce interventions) and at deployment for safety-shielded control.}
    \label{fig:overview}
\end{figure*}

\subsection{Safe Reinforcement Learning with a Predictive Safety Critic}

We consider a standard continuous-control reinforcement learning problem with
state $s_t \in \mathbb{R}^n$ and action $u_t \in \mathcal{U}\subset\mathbb{R}^m$.
A stochastic policy $\pi_\theta(u_t\mid s_t)$ with parameters $\theta$ generates
a nominal action $u_t^{\text{0}} = \pi_\theta(s_t)$.
We apply a safety shield that minimally modifies the raw action to a safe action
$\tilde u_t = T(x_t, u_t^{\text{0}})$, and the environment evolves according to system dynamics
$s_{t+1} \sim p(s_{t+1}\mid s_t, \tilde u_t)$.
Here, we distinguish the policy state $s_t$ from the safety critic input $x_t$ used by the critic and the safety filter.
The vector $x_t$ consists of safety-relevant inputs (typically a subset or transformation of $s_t$).
This notation allows us to present both a state-based safety critic and a state--action safety critic under a unified interface, and introduce them separately later.

The RL objective is to maximize the expected discounted return under safety constraints.
\begin{equation}
\begin{aligned}
    \max_{\theta}\;&
    \mathbb{E}_{\pi_\theta}\!\Big[\sum_{t=0}^{T-1}\gamma^t r(s_t,u_t)\Big] \\
    \text{s.t. }\;&
    s_{t+i} \in \mathcal{C},\quad \forall\,t=0,\dots,T-1,\ \forall\,i=0,\dots,H,
\end{aligned}
\label{eq:rl-objective}
\end{equation}
where $\mathcal{C}\subset\mathbb{R}^n$ denotes the safe set, and H denotes the length of the horizon.

To get the safety shield $T$, instead of hand-designing constraints in the state space, we introduce a
\emph{predictive safety critic} $h_\phi(x_t)$, parameterized by $\phi$. We may omit $\phi$ later for brevity. 
This scalar function maps a state $x_t$ to a safety value whose sign and magnitude
encode whether the system is expected to remain safe over the horizon of length $H$.
We interpret the \emph{safe set} as
\[
    \mathcal{C}_{\gamma_c} = \{x_t \mid h_\phi(x_t) \ge \gamma_c\},
\]
where $\gamma_c > 0$ is a conservative safety margin.
The overall goal is therefore to learn
\begin{itemize}
    \item a policy $\pi_\theta$ that optimizes \eqref{eq:rl-objective},
    \item a parameterized safety critic $h_\phi$ that predicts horizon-wise safety and admits
          a barrier-like interpretation,
    \item a differentiable safety filter $T$ based on $h_\phi$ that modifies $u_t^{\text{0}}$ into
          a safe action $\tilde u_t$ consistent with the safety critic.
\end{itemize}

\subsection{Predictive Receding-Horizon Safety Critic}

We now describe how the predictive safety critic $h_\phi(x)$ is trained
from data and how its training objective is structured.

\paragraph{Data and horizon-wise labeling.}
During interaction, we collect trajectories $\tau = \{(s_t, u_t, r_t, s_{t+1})\}_{t=0}^{T-1}$ under an exploratory or partially trained policy.
From these trajectories we derive binary labels $y_t \in \{0,1\}$ that indicate
whether the system remains non-terminated over the next $H$ steps:
\[
y_t =
\begin{cases}
1, & \text{if there is no terminal event in } (t, t+H],\\[3pt]
0, & \text{otherwise}.
\end{cases}
\]
All such samples are stored in off-policy replay buffer $\mathcal{D}$ used for training $h_\phi$.

\paragraph{Horizon-safety classification loss.}
We interpret the safety critic as predicting the probability of remaining safe
over the next $H$ steps. Specifically, we model
\[
    p(y_t=1 \mid s_t) \approx \sigma\!\big(h_\phi(x_t)\big),
\]
where $\sigma(\cdot)$ is the sigmoid function.
To account for the fact that negative labels induced by an eventual constraint
violation are typically more reliable when the violation is temporally closer, we
introduce an exponentially decaying discount weight for unsafe samples.
Let $n_t\in\{0,1,\dots,H\}$ denote the number of steps from time $t$ until the first
constraint violation within the horizon (only relevant when $y_t=0$), and let
$\gamma_{\text{label}}\in(0,1]$ be a discount factor. We define
\[
    w_{\text{label},t} \;=\;
    \begin{cases}
        1, & y_t = 1,\\[2pt]
        \gamma_{\text{label}}^{\,n_t}, & y_t = 0,
    \end{cases}
\]
so that unsafe samples with imminent violations receive higher weight,
while unsafe samples whose violations occur far in the future are down-weighted due
to increased uncertainty. For horizon-safe samples ($y_t=1$), no analogous
time-to-event notion is required because the label corresponds to remaining safe over
the entire horizon.
The resulting weighted binary cross-entropy loss is
\begin{equation}
\label{eq:loss-bce}
\begin{aligned}
    \mathcal{L}_{\text{label}}(\phi)
    = -\,\mathbb{E}
      \!\Big[
    \, w_{\text{label},t}\,
        \big(
        y_t\,\log\sigma(h_\phi(x_t) )
        \\ + 
        (1-y_t)\,\log(1-\sigma(h_\phi(x_t)))
        \big)
      \Big].
\end{aligned}
\end{equation}

\paragraph{Temporal safety-consistency loss.}
To further structure $h_\phi(x)$ beyond pure classification, we impose a
\emph{temporal consistency} regularizer inspired by discrete-time control barrier
functions (DT-CBFs). For safe transitions, a forward-invariance-style condition encourages
the safety value not to decrease, up to a class-$\mathcal{K}$ term
$\tilde{\alpha}$ (e.g., $\tilde{\alpha}(s)=\eta_{\mathrm{CBF}} s$):
\begin{equation}
\label{eq:dt_cbf}
h(x_{t+1})-h(x_t) + \tilde{\alpha}\!\big(h(x_t)-\gamma_c\big)\ge 0.
\end{equation}

For unsafe states ($y_t=0$), it is often more meaningful to encourage
progress away from the decision boundary: when a state is known to be
unsafe, the next-step safety value should not increase and should decrease by a
margin that scales with the current violation $\gamma_c-h(x_t)$. We encode
these two directional constraints with hinge penalties. Denoting
$\Delta h_t := h(x_{t+1})-h(x_t)$ and $\mathrm{ReLU}(z)=\max(0,z)$, we
define the safe-transition loss
\begin{equation}
\label{eq:loss-temp-consistency-safe}
\begin{aligned}    
\mathcal{L}_{\mathrm{temp}}^{+}(\phi)
=
\mathbb{E}  \!\left[
\,\mathbb{I}[y_t=1]\,
\mathrm{ReLU}\!\Big(
-\Delta h_t - \tilde{\alpha}\big(h_\phi(x_t)-\gamma_c\big) 
\Big)
\right],
\end{aligned}
\end{equation}
and the unsafe-transition loss
\begin{equation}
\label{eq:loss-temp-consistency-unsafe}
\begin{aligned}
    \mathcal{L}_{\mathrm{temp}}^{-}(\phi)
 =
\mathbb{E} \!\left[
\,\mathbb{I}[y_t=0]\,
\mathrm{ReLU}\!\Big(
\Delta h_t - \tilde{\alpha}\big(\gamma_c-h_\phi(x_t)\big)
\Big)
\right].
\end{aligned}
\end{equation}

The overall temporal consistency regularizer is then
\begin{equation}
\label{eq:loss-temp-consistency}
\mathcal{L}_{\mathrm{temp}}(\phi)
=
\mathcal{L}_{\mathrm{temp}}^{+}(\phi)
+
\mathcal{L}_{\mathrm{temp}}^{-}(\phi).
\end{equation}

\paragraph{Regularization: gradient and weight penalties.}
To improve smoothness and robustness of the safety critic, we add two
regularizers.

First, we penalize large input gradients to control the local Lipschitz constant:
    $\mathcal{L}_{\text{lip}}(\phi)
    = \mathbb{E}_{x_t\sim\mathcal{D}}\!\big[
        \|\nabla_x h_\phi(x_t)\|_2^2
      \big]$.
This term stabilizes the critic and reduces sensitivity to small state
perturbations.

Second, we apply standard weight decay to discourage overly complex models:
   $ \mathcal{L}_{\text{wd}}(\phi)
    = \|\phi\|_2^2$.

Combining all components, the safety critic is trained by optimizing
\begin{equation}
\label{eq:ncbf-train}
\mathcal{L}(\phi) = 
 w_{\text{label}}\,\mathcal{L}_{\text{label}}
+ w_{\text{temp}}\,\mathcal{L}_{\text{temp}}
+ w_{\text{lip}}\,\mathcal{L}_{\text{lip}}
+ w_{\text{wd}}\,\mathcal{L}_{\text{wd}},
\end{equation}
where $w_{\text{label}}, w_{\text{temp}}, w_{\text{lip}}, w_{\text{wd}}$ weight
each term correspondingly.

\paragraph{Uncertainty-aware conservative safety prediction.}
\label{sec:ensemble}
A single deterministic classifier can be overconfident under distribution shift
or in sparsely covered regions of the state space. To capture uncertainty induced by finite data, function approximation, and the randomized domain parameters, we train
an ensemble of $M$ independently initialized predictors
$\{h\}_{i=1}^M$ on $\mathcal{D}$.
Each member produces a probabilistic horizon-safety prediction $h_{i}(x_t)$.
We then aggregate ensemble predictions using a Conditional Value-at-Risk (CVaR)
operator to obtain a conservative yet probabilistic safety estimate.

This aggregation emphasizes the lower tail (pessimistic members),
thereby discounting overconfident predictions. We report $\hat p_{\text{CVaR}}(x_t)$
as the safety critic's conservative probability of remaining safe over the next
$H$ steps, and use it at deployment as a conservative safety score for filtering. Furthermore, by estimating the epistemic uncertainty, the standard deviation $\sigma(x_t)$ can be used to control the strength of the safety filter, which we will discuss in Section~\ref{sec:safety-filter}.%

\paragraph{Data buffer and sampling.}
To stabilize safety-critic learning under a non-stationary policy and severe class
imbalance, we maintain two separate replay buffers: a \emph{safe buffer}
$\mathcal{D}^{+}$ that stores horizon-safe transitions and an
\emph{unsafe buffer} $\mathcal{D}^{-}$ that stores transitions that lead to (or
contain) safety violations. Using two buffers, rather than a
single mixed replay, follows common practice in safety learning and risk-aware RL
when negative events can be rare and their frequency changes over training \cite{long_term_safety_puze2024, neural_certificate_2023}.

This design is motivated by the fact that as the policy improves, unsafe events
become increasingly infrequent. Without explicit control of negative replay, the
training distribution for the safety predictor can drift toward predominantly
safe samples, causing the predictor to overfit to the current training stage,
forget previously observed failure modes, and degrade generalization under
distribution shift. By preserving a dedicated set of unsafe samples and
continuing to replay them, $\mathcal{D}^{-}$ acts as a long-term memory of rare
but safety-critical behaviors, helping maintain calibration and reduce false
negatives.

During training, each minibatch is constructed by sampling a fixed ratio from
the two buffers. Concretely, given minibatch size $B$ and positive sampling ratio
$\rho^+\in[0,1]$, we draw $B^{+}=\lfloor \rho^+ B \rfloor$ samples from $\mathcal{D}^{+}$
and $B^{-}=B-B^{+}$ samples from $\mathcal{D}^{-}$, and then merge them into a
single minibatch. The ratio
$\rho^+$ provides direct control over the effective class balance seen by the
safety predictor, allowing us to maintain sufficient exposure to unsafe samples
throughout training even when violations become rare.

Optimizing \eqref{eq:ncbf-train} by combining from all components explained in this section yields a predictive safety critic $h_\phi(x)$
that (i) predicts the probability of remaining safe over the next $H$ steps conservativey and being aware of the uncertainty,
and (ii) is shaped to behave like a scalar barrier, which we can exploit to
design a safety filter.

\subsection{Differentiable Safety Shield via Closed-Form Projection}
\label{sec:safety-filter}
We now describe how to use $h_\phi$ to construct a differentiable safety filter
that maps the nominal action $u_t^{\text{0}}$ to a corrected action
$\tilde u_t$ that satisfies barrier-like constraints derived from the DT-CBF condition in \eqref{eq:dt_cbf}:
\begin{equation}
\label{eq:dt_cbf_rep}
h(x_{t+1}(u_t))-h(x_t) + \tilde{\alpha}\!\big(h(x_t)-\gamma_c\big)\ge 0.
\end{equation}
The key design choice is how to represent the next-step safety term $h(x_{t+1})$ inside \eqref{eq:dt_cbf_rep}. We consider two variants:
(i) a \emph{state} critic $h_\phi(x)$ with state-only input $x_t=f(s_t)$ combined with a differentiable one-step transition model, and
(ii) a \emph{state–action} critic $h^q_\phi(s_t, u_t)$ with state and action inputs that directly predicts the next-step safety effect of an action. The difference is analogous to the value function and the state action value function in reinforcement learning. Both lead to an affine constraint of the form $a^\top u \ge c$, enabling the subsequent unified closed-form projection. 

\paragraph{State Safety Critic}
In the state-critic variant, we obtain the next-step safety value $h_\phi(x_{t+1}(u))$ by composing a differentiable one-step model
\[
    x_{t+1}(u) = \hat{f}_\psi(x_t,u),
\]
with the safety critic. In our experiments, we directly use the differentiable nominal MJX simulator dynamics $\hat f_{\text{nom}}$.

To formulate a per-step safety projection, we linearize $h_\phi(x_{t+1}(u))$ with respect to $u$ around the nominal action $u_t^{\text{0}}$:
\begin{equation}
\label{eq:linearization-h}
\begin{aligned}
    h_\phi\big(x_{t+1}(u)\big)
    &\approx h_\phi\big(x_{t+1}(u_t^{\text{0}})\big)
    + \nabla_u h_\phi\big(x_{t+1}(u)\big)\big|_{u_t^{\text{0}}}^\top (u - u_t^{\text{0}}).
\end{aligned}
\end{equation}
We define
$
    a^h_t \triangleq \nabla_u h_\phi\big(x_{t+1}(u)\big)\big|_{u_t^{\text{0}}},
$
which is obtained by automatic differentiation through $\hat f_\psi$ (in MJX) and $h_\phi$.

\textbf{Constraint formulation.}
Substituting \eqref{eq:linearization-h} into \eqref{eq:dt_cbf_rep} yields an affine constraint in $u$:
\[
    (a^h_t)^\top u \;\ge\; c^h_t, \text{with}
\]
$    c^h_t \triangleq
    (a^h_t)^\top u_t^{\text{0}}
    + h_\phi(x_t) - \tilde{\alpha}\!\big(h_\phi(x_t)-\gamma_c\big)
    - h_\phi\big(x_{t+1}(u_t^{\text{0}})\big)$.

A key advantage of this framework is that $a$ can be computed directly via
automatic differentiation through the nominal system dynamics model.
In simulation, this nominal model is already available: the MuJoCo/MJX
physics engine provides a differentiable mapping with no additional modeling assumptions. Leveraging this native differentiable transition model requires no extra learning, no auxiliary networks, and no system-identification phase. It injects known physics into the otherwise
model-free RL pipeline, yielding two major benefits:

\begin{itemize}
    \item Accurate action gradients. The term $\nabla_u h_\phi(x_{t+1})$
    is computed through the true simulated dynamics, providing more reliable
    sensitivity information than finite differences or black-box estimators.
    \item Zero additional computational burden.
    Since MJX already supports end-to-end differentiation, the linearization is
    obtained ``for free'' during backpropagation, without requiring a learned
    control-affine approximation.
\end{itemize}
But this method is prone to nominal model mismatch, which is crucial because of sim2real gaps. Under domain randomization or unmodeled effects, the executed dynamics $f_{\text{true}}$
can deviate from $\hat f_{\text{nom}}$, yielding biased rollouts
$\hat x_{t+1}\neq x_{t+1}$ and therefore biased estimates of both
$h_\phi(x_{t+1})$ and its action sensitivity. This mismatch can lead the safety
projection to correct actions in an inconsistent direction. 

\paragraph{State--Action Safety Critic}
To mitigate unavoidable model mismatch, we introduce a \emph{state--action} safety critic $h^q_\phi(x)$ with inputs $s_t,u_t$ and interpret it as the one-step safety value induced by applying $u$ at $s_t$, then we can infer the state safety critic value from the state-action safety value:
\begin{equation}
\label{eq:q-def}
    h(x_{t+1}(u_t)) \leftarrow h^q_\phi(s_t, u_t), \quad x_{t+1}\sim f_{\text{true}}(s_t,u).
\end{equation}
In practice, $h^q_\phi$ is trained from replayed transitions and horizon-safety labels, thereby absorbing the effective dynamics (including randomization) into the predictor.

Similar to \eqref{eq:linearization-h}, the linearization now becomes:
\begin{equation}
\label{eq:q-linearization}
\begin{aligned}
    h(x_{t+1}(u))
    &\approx h^q_\phi\!\big(s_t,u_t^{\text{0}}\big)
      + \nabla_u h^q_\phi\!\big(s_t,u\big)\big|_{u_t^{\text{0}}}^{\!\top}(u-u_t^{\text{0}}).
\end{aligned}
\end{equation}
We define
$
a^q_t \triangleq \nabla_u h^q_\phi\!\big(f(s_t,u)\big)\big|_{u_t^{\text{0}}},
$
which is obtained by differentiating through $h^q_\phi$ alone.

\textbf{Constraint formulation.}
For the state--action critic, the safety condition can be then directly written in terms of the executed action:
\[
    h^q_\phi(s_t,u) - h^q_\phi(s_{t-1},u_{t-1}) + \tilde{\alpha}\!\big(h^q_\phi(s_{t-1},u_{t-1})-\gamma_c\big) \ge 0,
\]
Substituting the linearization \eqref{eq:q-linearization} yields an affine constraint
\[
    (a^q_t)^\top u \;\ge\; c^q_t,
\]
with
$    c^q_t \triangleq
    (a^q_t)^\top u_t^{\text{0}}
    + (1-\tilde{\alpha})\,h^q_\phi(x_{t-1})
    - h^q_\phi\!\big(s_t,u_t^{\text{0}}\big)
    + \tilde{\alpha}\gamma_c.$

This construction avoids differentiating through a potentially mismatched
dynamics model: the gradient $a$ is obtained by automatic differentiation
through $q_\phi$ alone, while the training target for $q_\phi$ is grounded in
the realized transition data. As a result, the safety layer becomes more
computationally efficient and less sensitive to nominal-model errors introduced by domain randomization, but it can inherit uncertainty induced by randomization. 

\paragraph{Quadratic program with uncertainty-scaled slack}
Both variants above yield an affine constraint $(a^\cdot_t)^\top u \ge c^\cdot_t$. We enforce either constraint with the same one-constraint QP, differing only in whether $(a_t,c_t)=(a^h_t,c^h_t)$ or $(a_t,c_t)=(a^q_t,c^q_t)$.

Firstly, to allow controlled relaxation when uncertainty is high, we introduce a \emph{standard-deviation--scaled} slack variable $\xi_t \ge 0$ and enforce the relaxed constraint in standardized units:
\[
    a_t^\top u + \sigma_t \xi_t \;\ge\; c_t,
\]
so that the relaxation in the original units is $\varepsilon_t \triangleq \sigma_t \xi_t$. As defined in Section \ref{sec:ensemble}, $\sigma_t$ refers to the standard deviation of the safety critic ensemble. 

We then unify both formulations and compute a minimally modified action by solving the one-constraint QP
\begin{equation}
\label{eq:qp-simple-final}
\begin{aligned}
    \min_{u,\xi_t}\quad
        & \|u - u_t^{\text{0}}\|_2^2 + \lambda_s \xi_t^2 \\[2pt]
    \text{s.t.}\quad
        & a^\top u + \sigma_t \xi_t \;\ge\; c,\\
        & \xi_t \ge 0,
\end{aligned}
\end{equation}
where $\lambda_s>0$ penalizes the slack variable $\xi_t$.
This scaling makes the relaxation uncertainty-relevant: larger absolute constraint violation (in the original units) is permitted
only when the critic is more uncertain, while the penalty remains comparable across states in standardized units.

Because Eq.~\eqref{eq:qp-simple-final} has a single linear constraint, the KKT conditions yield a closed-form solution:
\begin{equation}
\label{eq:closed-form-qp-final}
\begin{aligned}
    u_t^\star
    &= u_t^{\text{0}}
       + \frac{\big[c - a^\top u_t^{\text{0}}\big]_+}
              {\|a\|_2^2 + \sigma_t^2/\lambda_s}\, a, \\[6pt]
    \xi_t^\star
    &=
    \frac{\sigma_t\,\big[c - a^\top u_t^{\text{0}}\big]_+}
         {\lambda_s\|a\|_2^2 + \sigma_t^2},
    \qquad
    [z]_+ \triangleq \max\{z,0\}.
\end{aligned}
\end{equation}
If $a^\top u_t^{\text{0}} \ge c$, the constraint is inactive and the action is left unchanged, $u_t^\star = u_t^{\text{0}}$.
Otherwise, the safety filter moves the action along the steepest-ascent direction $a$
by the smallest amount needed to satisfy the constraint, while $\xi_t^\star$ provides controlled relaxation when strict feasibility is not possible.
We define the \emph{safety-filtered action} as
\[
    \tilde u_t
    = T_\phi(x_t, u_t^{\text{0}}).
\]

The mapping $T_\phi(x_t, u_t^{\text{0}})$ is end-to-end differentiable with respect to the policy parameters $\theta$ via $u_t^{\text{0}}$ and the differentiable closed form solution of the QP, and does not require differentiating through a dynamics model at deployment.

\subsection{Safe Policy Learning with Anticipation Loss}

We now describe how we integrate the safety critic and safety filter into
policy optimization.

During rollouts, we generate nominal actions
$u_t^{\text{0}} \sim \pi_\theta(\cdot\mid s_t)$ and then apply the safety
filter to obtain $\tilde u_t = T_\phi(x_t, u_t^{\text{0}})$. If the policy naively ignores safety, the filter will frequently intervene,
leading to a mismatch between the actions used to compute policy gradients
and those actually executed. Meanwhile, if the safe action deviates a lot from the raw action, the linearization error in Eq.\ref{eq:linearization-h} or \ref{eq:q-linearization} becomes larger which might lead in inaccurate safety modifications.  
To mitigate this, we introduce an \emph{anticipation loss} that explicitly
encourages the policy to output actions close to the
safety-filtered actions:
\begin{equation}
\label{eq:anticipate-loss}
\mathcal{L}_{\text{anticipate}}(\theta)
= \mathbb{E}
  \Big[
    \big\|
      \mu_\theta(s_t)
      - T_\phi(x_t, u^0_t)
    \big\|_2^2
  \Big].
\end{equation}
coefficient.
Gradients of \eqref{eq:anticipate-loss} are backpropagated through
$T_\phi$, thereby aligning the policy with the safety filter and reducing
the frequency and magnitude of interventions.

We employ Proximal Policy Optimization (PPO) \cite{schulman_proximal_2017} as the base RL algorithm.
Let $\hat{A}_t$ denote generalized advantage estimates.
The PPO surrogate objective is
\begin{equation}
\label{eq:ppo-loss}
\begin{aligned}
    \mathcal{L}_{\text{PPO}}(\theta)
    &= \mathbb{E}_t\!\left[
        \min\!\Big(
          r_t(\theta)\,\hat{A}_t,\;
          \mathrm{clip}\big(r_t(\theta),\,1-\epsilon,\,1+\epsilon\big)\,\hat{A}_t
        \Big)
      \right], \\
    r_t(\theta)
    &= \frac{\pi_\theta(u_t\mid s_t)}{\pi_{\theta_{\text{old}}}(u_t\mid s_t)},
\end{aligned}
\end{equation}
where $\epsilon$ is the PPO clipping parameter.
We jointly optimize the policy $\pi_\theta$ and value function $V_\psi$ via
the total loss
\begin{equation}
\label{eq:total-loss-biased}
\begin{aligned}
\mathcal{L}_{\text{total}}(\theta,\psi)
= & -\,\mathcal{L}_{\text{PPO}}(\theta) 
 + \lambda_v\,\mathbb{E}_t\!\big[(V_\psi(s_t)-\hat{R}_t)^2\big] \\
  & - \lambda_H\,\mathbb{E}_t\!\big[\mathcal{H}(\pi_\theta(\cdot\mid s_t))\big] \\
  & + \lambda_{\text{anticipate}}\,
      \mathcal{L}_{\text{anticipate}}(\theta),
\end{aligned}
\end{equation}
where $\mathcal{H}(\pi_\theta)$ is the policy entropy and
$\lambda_v,\lambda_H,\lambda_{\text{anticipate}}$ are scalar weights.

The training procedure alternates between rollout with shielded policy, updating the safety critic and update the policy, as presented in Algorithm~\ref{alg:safe-rl-safety-critic}.

\begin{algorithm}[t]
\caption{Safe RL with Predictive Safety Critic and PPO}
\label{alg:safe-rl-safety-critic}
\begin{algorithmic}[1]
\State Initialize policy $\pi_\theta$, value function $V_\psi$,
       predictive safety critic $q_\phi$.
\State Initialize RL replay buffer and safety-critic buffer $\mathcal{D}$.
\While{not converged}
    \State \textbf{Rollout with safety filter:}
    \For{environment step $t = 0,\dots,T-1$}
        \State Sample nominal action $u_t^{\text{0}} \sim \pi_\theta(\cdot\mid s_t)$.
        \State $\tilde u_t = T_\phi(x_t, u_t^{\text{0}})$ via \eqref{eq:closed-form-qp-final} and execute.
        \State Store $(s_t, u_t^{\text{0}}, \tilde u_t, r_t, s_{t+1}, \text{done}_t)$ in RL buffer.
    \EndFor
    \State Construct horizon labels $y_t$ from rollouts and
           append $(s_t, s_{t+1}, y_t)$ to $\mathcal{D}_+$ and $\mathcal{D}_+$.

    \State Update safety critic $q_\phi$ using \eqref{eq:ncbf-train}.
    \State Update $\pi_\theta$ and $V_\psi$ using \eqref{eq:ppo-loss}.
\EndWhile
\end{algorithmic}
\end{algorithm}

\begin{figure*}[tbh]
\captionsetup{skip=2pt}

\centering
    \includegraphics[width=\textwidth]{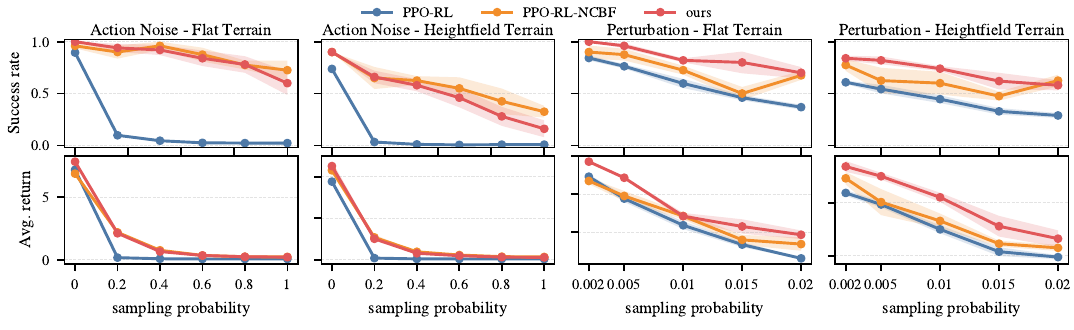}
    \caption{Top row: success rate; bottom row: average return. Columns correspond to four evaluation suites (action-noise and external-perturbation, each on flat terrain and heightfield terrain). Curves show mean performance over independent runs for PPO-RL, PPO-RL-NCBF, and our method, with shaded bands indicating standard error across different seeds. The x-axis denotes the perturbation/noise sampling probability.}
    \label{fig:exp_all}
\end{figure*}

\section{Experiments and Results}

We evaluate our predictive safety critic and differentiable safety shield through a sequence of experiments that isolate (i) safety prediction quality, (ii) closed-loop safety and performance under distribution shift, and (iii) key design choices. All policies are trained in simulation using MJX with GPU acceleration on a Unitree Go2 model.

As baselines, we compare against state-of-the-art PPO-style locomotion pipelines equipped with standard domain randomization and common stabilization tricks (details in the appendix), and with a neural control barrier function. To test safety generalization beyond the training distribution, we introduce controlled sim-to-sim shifts between training and evaluation distributions, including terrain changes (flat $\rightarrow$ uneven), action noise, observation noise, and external perturbations. This suite is designed to emulate the failure modes that dominate legged locomotion and to expose cases where myopic, single-step safety heuristics break down.

We organize the results as follows. First, we validate the safety critic under the above shifts, demonstrating that it anticipates imminent failures over a receding horizon and that the ensemble uncertainty increases in out-of-distribution regimes. Second, we evaluate the full system (critic + closed-form projection) and report success/failure rates and returns under increasing shift and perturbation strength, comparing to baselines with and without safety mechanisms, and over baselines with neural control barrier functions. Finally, we perform an ablation over the prediction horizon $H$ to quantify the trade-off between early risk detection, intervention frequency, and locomotion performance.

\subsection{Safety prediction}
\label{sec:result:safety_critic}

In this section, we validate the predictive safety critic using the state–action formulation, which is less sensitive to transition-model mismatch and empirically generalizes better under domain shifts; results for the state-only formulation are deferred to Appendix~\ref{app: state model}. The critic is trained on flat terrain with domain randomization, and evaluated under a shifted test distribution with randomly generated uneven terrains and enlarged external perturbations. As intended, the ensemble produces conservative safety estimates, with the CVaR aggregate dropping ahead of failures and uncertainty increasing in out-of-distribution regions. We further visualize a representative rollout to show when the safety condition becomes active and how the closed-form shield modifies the policy actions in response.

\begin{figure}[tbh]
    \centering
    \includegraphics[width=\columnwidth]{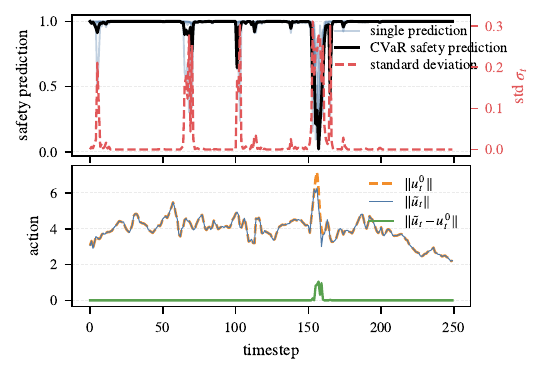}
    \caption{Example rollout diagnostics of the predictive safety shield.
\emph{Top:} ensemble safety predictions (gray, ``single prediction''), their CVaR-aggregated safety estimate $\hat p_{\mathrm{CVaR}}(x_t)$ (black), and the ensemble standard deviation $\sigma_t$ (red, right axis). Peaks in $\sigma_t$ indicate increased epistemic uncertainty and coincide with sharp drops in the predicted safety.
\emph{Bottom:} action magnitudes before and after shielding, showing the raw action norm $\|u_t^{0}\|$ (orange), the shielded action norm $\|\tilde u_t\|$ (blue), and the correction magnitude $\|\tilde u_t-u_t^{0}\|$ (green). Non-zero corrections occur primarily during high-uncertainty and low-safety intervals, illustrating successful interventions.}
    \label{fig:exp_pred}
\end{figure}

\subsection{General Performance}
\label{subsec:general_performance}
We next evaluate closed-loop robustness under controlled sim-to-sim distribution shifts, focusing on two failure modes that commonly break legged locomotion controllers: stochastic action noise, which simulates a suboptimal policy, and impulsive external disturbances. We test each shift on two terrain families (flat and heightfield), resulting in four evaluation settings. For each setting, we evaluate different rates at which the noise and perturbations occur, and report both success rate and average return (Fig.~\ref{fig:exp_all}). More detailed explanations are presented in Appendix \ref{app: domain_randomization}

We compare three methods: (i) PPO-RL, a PPO locomotion baseline trained with standard domain randomization\cite{sabouni_reinforcement_2025}; (ii) PPO-RL-NCBF, the same baseline augmented with a neural CBF safety shield; and (iii) ours, our predictive safety critic with a differentiable closed-form safety projection.

\textbf{Action noise.}
PPO-RL is highly brittle under action noise: success collapses rapidly even at low noise rates and returns drop to near-zero across both terrains. In contrast, both safety-equipped methods maintain substantially higher success, with our approach consistently exhibiting the strongest robustness as the noise becomes more frequent. The improvement is most pronounced at moderate-to-high noise probabilities, where the baseline fails almost deterministically while our method still completes a large fraction of episodes and retains non-trivial returns. This gap increases on uneven terrain, indicating that action noise combined with terrain-induced contact variability exacerbates failures of the unshielded policy.

\textbf{External perturbations.}
Under impulsive perturbations, performance degrades more smoothly for all methods as the perturbation frequency increases, but the ordering remains consistent. PPO-RL shows the steepest drop in both success and return, particularly on heightfields. Adding an NCBF shield improves robustness over PPO-RL, but our method achieves the best overall safety-performance trade-off: it maintains the highest success rates across the sweep and preserves higher returns, suggesting that the predictive critic is able to anticipate destabilizing events and trigger earlier, less disruptive interventions than single-step safety heuristics.

Overall, these results demonstrate that our method, incorporating predictive risk estimation with a differentiable projection, yields markedly better safety generalization under distribution shift. The gains are strongest in regimes where disturbances occur frequently and where terrain irregularity amplifies transient errors, which are exactly the conditions that dominate real-world failure cases in legged locomotion.

\subsection{Ablation on Horizon Length}
\begin{figure}
    \centering
    \includegraphics[width=\columnwidth]{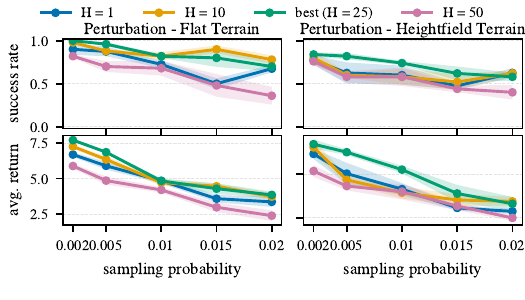}
    \caption{Ablation on prediction horizon H for selecting the safety-critic/shield horizon under sim-to-sim domain shift. Experiment settings are similar to Fig.~\ref{fig:exp_all}. Curves show mean performance, with shaded bands indicating standard error across different seeds}
    \label{fig:exp_H}
\end{figure}

Fig.~\ref{fig:exp_H} reports an ablation on the prediction horizon $H$ of the receding-horizon safety critic under the domain shift with enlarged external perturbations, evaluated on both flat and heightfield (uneven) terrains, with the model trained on flat terrain. Using similar notation as in the previous section, this experiment isolates how far-horizon safety prediction impacts both closed-loop robustness and locomotion performance when the evaluation dynamics deviate from training.

A key motivation for the receding-horizon critic is to address the myopia of classical and neural CBFs, where single-step or local constraints can fail to anticipate failures several steps in advance. Increasing $H$ extends the look-ahead window and improves early risk detection, which tends to raise success under stronger perturbations---especially on heightfields where contacts and footholds introduce longer-term instabilities.

However, the figure also highlights a clear bias–variance / generalization trade-off. As $H$ grows, training becomes harder (longer credit assignment, noisier supervision, and a larger effective hypothesis class). In practice, longer-horizon critics are more prone to overfitting to the training distribution and to producing overly conservative risk estimates under shift. This conservatism increases intervention frequency and reduces returns, and can even lower success when the shield becomes too restrictive. Consistent with this, the $H=50$ setting tends to degrade performance relative to moderate horizons, particularly as perturbations intensify and the heightfield dynamics differ from those seen during training.

Overall, $H=25$, which corresponds to 0.5 seconds, provides the best compromise under domain shift: it is long enough to mitigate myopic failure modes (improving robustness at higher perturbation strengths), yet not so long that the critic becomes difficult to train or excessively conservative, preserving higher average return. The complete results, also under action noise, are presented in Appendix~\ref{app:ablation_H}. 

\section{Conclusion}
We presented a safe locomotion framework that replaces myopic one-step safety constraints with a receding-horizon safety critic and a closed-form, differentiable safety shield. The critic predicts horizon-wise feasibility from rollout-derived labels and is stabilized through temporal-consistency and smoothness regularization, while an ensemble with CVaR aggregation provides conservative safety estimates under distribution shift. The resulting safety layer performs minimal action corrections without requiring a dynamics model at deployment and can be trained end-to-end with PPO via a safety-anticipation loss, reducing policy--filter mismatch.

Across simulated quadruped locomotion under controlled domain shifts (action noise, external perturbations, and terrain changes), our method improves success rates while preserving competitive returns relative to PPO with domain randomization and an NCBF-based shield. An ablation over the horizon length highlights a practical trade-off between anticipatory safety and critic trainability/conservatism.

A promising direction for future work is to fine-tune the safety critic using data from real world experiments to further enhance the robustness when faced with sim2real gap.

\bibliography{main}
\bibliographystyle{plainnat}

\appendix
Appendices are submitted as supplemental materials.
Our codebase will be released after the double-blind review phase. 

\subsection{Hyperparameters}
\label{appendix:hyperparameter}
In this section, we list the most relevant hyperparameters selected for our method. For the complete set of hyperparameters and their values, we refer to our codebase, which will be released after acceptance.

\begin{table}[tbh]
\centering
\small
\setlength{\tabcolsep}{8pt}
\renewcommand{\arraystretch}{1.1}
\caption{Hyperparameters.}
\label{tab:hyperparams}
\begin{tabular}{@{}lll@{}}
\toprule
\textbf{Hyperparameter} & \textbf{Symbol} & \textbf{Value} \\
\midrule
Safety prediction horizon & $H$ & 5 \\
Safety margin (threshold) & $\gamma_c$ & 0.5 \\
Unsafe-label discount & $\gamma_{\text{label}}$ & 0.95 \\
CVaR tail probability & $\alpha_{\text{cvar}}$ & 0.4 \\
Ensemble size & $M$ & 5 \\
CBF gain (for $\tilde{\alpha}(s)=\eta_{\text{CBF}}s$) & $\eta_{\text{CBF}}$ & 0.5 \\
Sampling ratio from safe buffer & $\rho^+$ & 0.2 \\
Loss weight: horizon-safety classification & $w_{\text{label}}$ & 1 \\
Loss weight: temporal consistency & $w_{\text{cbf}}$ & 0.5 \\
Loss weight: Lipschitz/gradient penalty & $w_{\text{lip}}$ & 0.0001 \\
Loss weight: weight decay & $w_{\text{wd}}$ & 0.00001 \\
QP slack penalty & $\lambda_s$ & 1000 \\
Anticipation loss weight & $\lambda_{\text{anticipate}}$ & 0.01 \\
\bottomrule
\end{tabular}
\end{table}

\subsection{Two Variants of the Safety Critic}
\label{app: state model}

\begin{figure*}[t]
    \centering
    \includegraphics[width=\textwidth]{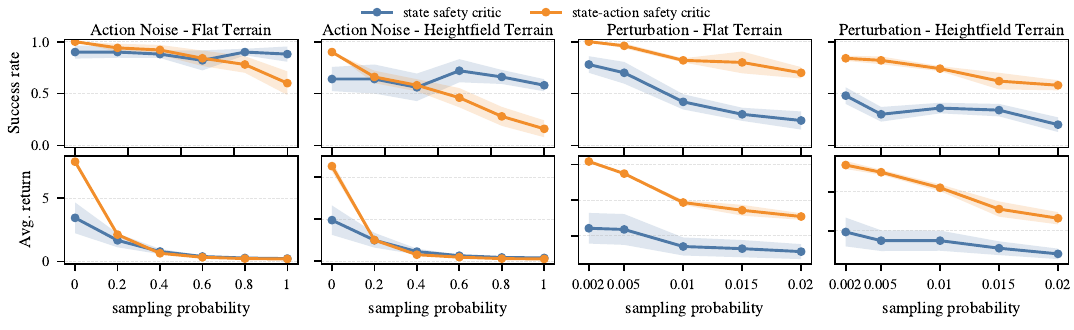}
    \caption{Comparison of state-only and state-action safety critics. The state-action safety critic demonstrates better generalizability and robustness, particularly under perturbations and domain shifts. The results are shown for success rate (top) and average return (bottom) across different perturbation settings. The result is aggregated over 5 seeds, and the shaded area shows the standard error. The state-action model outperforms the state-only model under external perturbation and domain shift.}
    \label{fig:app:exps_h_vs_q}
\end{figure*}

We introduce two variants of the safety critic in the main paper: (i) the state-only safety critic, which relies on a nominal dynamics model to predict safety, and (ii) the state--action safety critic, which integrates both the state and the action to provide a more accurate safety estimate. 

\begin{figure}
    \centering
    \includegraphics[width=\linewidth]{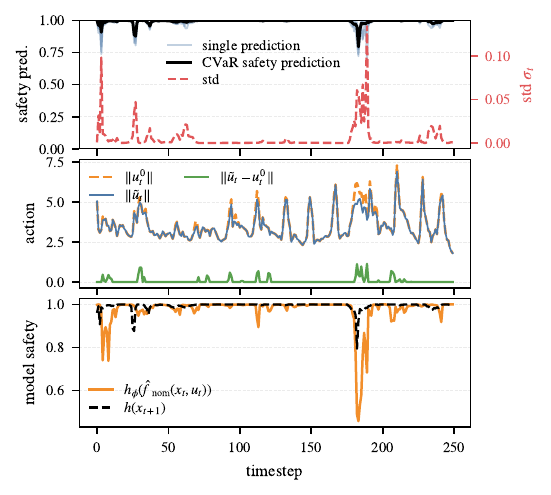}
    \caption{Prediction of the state-based safety critic under domain shift.
    \emph{Top:} ensemble safety predictions (gray), CVaR-aggregated safety prediction (black), and ensemble standard deviation $\sigma_t$ (red dashed, right axis).
    \emph{Middle:} action norms before and after shielding, $\|u_t^0\|$ (orange dashed), $\|\tilde u_t\|$ (blue), and correction magnitude $\|\tilde u_t-u_t^0\|$ (green).
    \emph{Bottom:} nominal-model-based one-step safety estimate $h_\phi(\hat f_{\mathrm{nom}}(x_t,u_t))$ (orange) versus the corresponding value evaluated on the realized next state $h(x_{t+1})$ (black dashed).
    Under heightfield terrain with enlarged perturbations, the nominal estimate can diverge substantially from the next-state reference (e.g., around $t\!\approx\!175$--$190$), leading to spurious risk alarms and unnecessary interventions.}
    \label{fig:app:state_model_prediction}
\end{figure}

Under nominal model mismatch, a model-based safety layer can compute an incorrect one-step lookahead and thus fail to map the raw action into the true safe-action set. This misalignment can degrade the robustness of closed-loop performance using a safety layer based on the state-only safety critic, especially under large distribution shifts. To evaluate these models, we use the same evaluation protocol as in the main paper and visualize the predicted safety signals and the resulting action modifications in Fig.~\ref{fig:app:state_model_prediction}. We focus on the most challenging setting—heightfield terrain with enlarged external perturbations—which induces the largest sim-to-sim shift.

The results shown in Fig.~\ref{fig:app:state_model_prediction} highlight a key failure mode for state-model-based shielding. In the third panel, the nominal-model-based safety estimate $h_\phi(\hat f_{\mathrm{nom}}(x_t, u_t))$ (orange) deviates significantly from the ground truth safety value evaluated on the actual next state $h(x_{t+1})$ (black dashed). The mismatch is particularly noticeable around $t\approx175$--$190$, where the nominal model predicts a sharp drop in safety (down to $\sim 0.5$), even though the actual next-state safety value remains close to 1 with only a mild decrease. This erroneous pessimism correlates with spikes in epistemic uncertainty (top panel, red dashed $\sigma_t$) and triggers unnecessary action corrections by the safety layer (middle panel, green), even though the next-state reference suggests substantially lower risk. These results indicate that, under large domain shifts, the nominal model can distort the safety estimate and cause the safety shield to intervene at inappropriate times, either by being overly conservative or by missing imminent failures when the mismatch flips sign. This limitation motivates the use of state--action critics that do not rely on a potentially miscalibrated nominal dynamics model.

In terms of generalizability and robustness, the state-action variant outperforms the state-only one, particularly in handling domain shifts, as shown by the experiment in Fig.~\ref{fig:app:exps_h_vs_q}, which is similar to those in the main paper. In the action noise test, the policies with the state-action model show slightly better results than those with the state-only safety critic at lower noise levels. However, as the action noise sampling ratio increases, the state-only model outperforms the state-action model. This is expected because the state-action model is more sensitive to large and frequent action noises due to the direct input of actions. When faced with perturbations and domain shifts, however, the safety layer based on the state-only safety critic and the nominal model struggles to capture the true dynamics because of the nominal model mismatch, leading to a clear performance gap in favor of the state-action model.

Because of the better generalizability and robustness of the state-action model, we choose to present it as our main method for the rest of the evaluations.

\subsection{Safety Critics vs. SafeFall Baseline}
\label{app: safefall}

A key contribution of our approach is a structured training pipeline for the safety critic, formulated as an unsafe-event predictor. In humanoid locomotion, falling is one of the most critical safety events and has been extensively studied in recent work. In particular, SafeFall~\cite{meng_safefall_2025} is closely related in spirit: it trains a predictor for the probability of falling, but differs in how supervision is constructed and how the loss is defined.

\begin{table*}[t]
\centering
\caption{Comparison of performance metrics between our state-action safety critic and SafeFall predictor under perturbations on flat and heightfield terrains.}
\begin{tabular}{|l|l|c|c|c|} 
\hline
\textbf{Metric} & \textbf{Experiment} & \textbf{State-Action Safety Critic} & \textbf{SafeFall} \\ 
\hline
Accuracy & Perturbation - Flat Terrain & 0.6160 $\pm$ 0.0200 & 0.6200 $\pm$ 0.0265 \\
Accuracy & Perturbation - Heightfield Terrain & 0.5800 $\pm$ 0.0254 & 0.5480 $\pm$ 0.0232 \\
False Positives & Perturbation - Flat Terrain & \textbf{0.0800} $\pm$ 0.0086 & 0.1680 $\pm$ 0.0171 \\
False Positives & Perturbation - Heightfield Terrain & \textbf{0.1760} $\pm$ 0.0197 & 0.2480 $\pm$ 0.0113 \\
False Negatives & Perturbation - Flat Terrain & 0.0736 $\pm$ 0.0053 & 0.0160 $\pm$ 0.0020 \\
False Negatives & Perturbation - Heightfield Terrain & 0.0612 $\pm$ 0.0071 & 0.0129 $\pm$ 0.0014 \\
Prediction Steps before Failure & Perturbation - Flat Terrain & \textbf{29.9833} $\pm$ 1.3152 & 12.0833 $\pm$ 0.4587 \\
Prediction Steps before Failure & Perturbation - Heightfield Terrain & \textbf{24.8333} $\pm$ 1.8499 & 16.8250 $\pm$ 0.5575 \\
\hline
\end{tabular}
\label{table:metrics_comparison}
\end{table*}

Our method trains the critic in a receding-horizon manner with temporally discounted labels, which yields \emph{dense} supervision at every timestep and supports flexible horizon selection at both training and deployment. In contrast, SafeFall discretizes each trajectory into three regions---safe ($t < T_1$), ambiguous ($T_1 < t < T_2$), and unsafe ($t > T - T_2$, where $T$ denotes the trajectory length)---using hand-crafted triggers tied to a fixed number of steps. This design encodes a similar intuition (early states are safer; states close to failure are riskier), but it imposes several practical limitations.

First, because the labeling is defined relative to a fixed trajectory length, it implicitly restricts training and data collection to rollouts with a pre-specified horizon. This can reduce label coverage when episodes terminate early or vary in length. Additionally, by construction, each trajectory is anchored around a single terminal event, which limits the representation of repeated near-failure episodes or multiple failure modes within the same rollout (e.g., stumble--recovery--fall). Second, the supervision is highly sensitive to the threshold choices $(T_1, T_2)$: shifting either boundary changes which states are treated as safe versus ambiguous/unsafe, directly affecting both class balance and the sharpness of the decision boundary learned by the predictor. As a result, the predictor quality can degrade under distribution shift if the step-based segmentation no longer aligns with the true time-to-failure statistics. Third, our safety predictor leverages an ensemble of models to better capture uncertainty. The ensemble approach allows the safety critic to provide more accurate risk estimates and provide more conservative predictions, which is particularly valuable when faced with the complexities of real-world environments.

In this section, we compare the SafeFall-style falling predictor with our receding-horizon safety predictor along three key dimensions: (i) prediction quality under domain shift through visualization, (ii) quantitative evaluation of the prediction quality, and (iii) closed-loop performance, where each predictor is used to instantiate a safety layer. Performance is measured by success rate and average return.

For SafeFall, we adopt the parameterization from the original paper, setting $T_1 = \frac{2T}{3}$ and $T_2 = T - 100\,\mathrm{ms}$. In our implementation, rollouts are truncated to a fixed horizon of $T=10\,\mathrm{s}$ and simulated at 50\,Hz.

\begin{figure}
    \centering
    \includegraphics[width=\columnwidth]{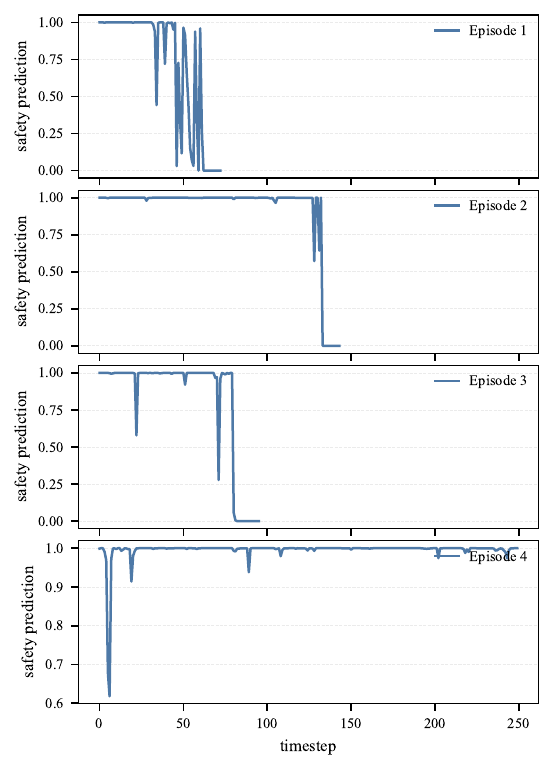}
\caption{Safety predictions from the SafeFall predictor across several random episodes. The trajectory length is set to 250 timesteps. SafeFall's predictions tend to be more binary, predicting failure only shortly before the event. This contrasts with our receding-horizon method, which provides earlier and more gradual safety estimates.}
    \label{fig:app:safefall_prediction}
\end{figure}

\paragraph*{Visualization of the Predictions}
In Fig.~\ref{fig:app:safefall_prediction}, we present safety predictions from the SafeFall predictor across several random episodes. The trajectory length is set to 250 timesteps. As shown, the SafeFall predictor tends to provide more binary predictions, where the safety value sharply drops right before the failure event occurs. This characteristic of the SafeFall predictor means that it is unable to provide sufficient early warning of an impending failure, limiting the time available for the safety layer to intervene. In contrast, our method offers more gradual safety predictions, which allows it to identify potential failure risks earlier in the episode.

SafeFall uses the predictor as a switch to turn on a backup policy and, therefore, has less stringent requirements on reaction time and gradient. However, the ability to predict failures robustly and as early as possible is crucial for a safety layer. By identifying the risk of failure sooner, the system can take corrective actions proactively, preventing the system from reaching an unrecoverable state. Furthermore, the softer gradients in the safety prediction lead to more gradual adjustments in the action, preventing large, abrupt changes that could exceed action limits or cause instability in the system. This smoother action modification is essential for maintaining continuous and safe operation, as sudden jumps in action could result in undesirable system behavior or damage. Therefore, our approach not only offers earlier failure detection but also produces safer, more controlled interventions compared to the binary predictions of the SafeFall-style predictor.

\begin{figure*}[t]
    \centering
    \includegraphics[width=\textwidth]{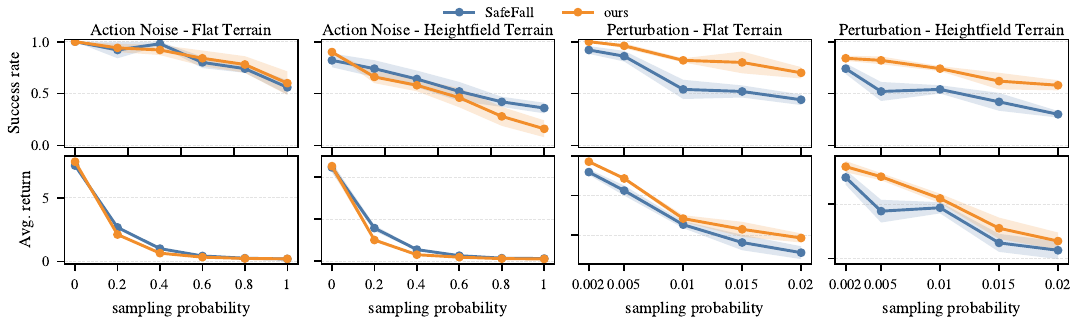}
\caption{Comparison of SafeFall and our method in Safe RL training using each as a safety critic. The plot compares success rate (top row) and average return (bottom row) across different levels of action noise and external perturbations. Both methods show similar robustness under action noise, but our method outperforms SafeFall when faced with perturbations, demonstrating its superior generalizability.}
    \label{fig:app:safefall_exps}
\end{figure*}

\paragraph*{Quantitative Evaluation}
Additionally, we compute several metrics to evaluate prediction accuracy:
\begin{itemize}
    \item Accuracy (how often unsafe and safe episodes are predicted correctly),
    \item False positive rate (the proportion of unsafe episodes predicted as safe),
    \item False negative rate (the proportion of safe episodes predicted as unsafe),
    \item Average prediction steps before failure (the number of steps before an unsafe episode when the predictor is able to predict the failure).
\end{itemize}

Since SafeFall has by design a shorter predictions horizon, we calculate accuracy, false positive rate and false negative rate based on the last five steps only to have a fail comparison. To evaluate the ability of predicting failure in advance, we introduce the last metric, indicating how many steps in advance is the predictor able to predict possible failures. 

The values reported in Table \ref{table:metrics_comparison} are aggregated over all seeds, episodes, and sampling ratios of the perturbation. As shown in the table, the state-action safety critic demonstrates similar accuracy to the SafeFall predictor in most cases. However, the SafeFall predictor exhibits a higher false positive rate, particularly under perturbations on heightfield terrain, highlighting the effectiveness of our uncertainty-aware, conservative predictions in handling domain shifts. While our method does result in a higher false negative rate, we argue that a more conservative approach, prioritizing safety by avoiding false positives, is crucial in high-risk scenarios. A higher false negative rate ensures that potential failures are not prematurely classified as safe, allowing for timely interventions before reaching unsafe states.

The prediction steps for true positives further demonstrate the advantage of the state-action safety critic. On flat terrain, it predicts failures an average of 29.98 steps ahead, while the SafeFall predictor identifies failures only 12.08 steps ahead. On heightfield terrain, the state-action model predicts failure at 24.83 $\pm$ 1.85 steps, compared to 16.83 $\pm$ 0.56 for SafeFall. This ability to predict failures earlier is crucial for proactive safety interventions and reinforces the generalizability of our method in complex environments.

\paragraph*{Closed-Loop Performance}
In Fig.~\ref{fig:app:safefall_exps}, we compare the performance of SafeFall and our method in the context of Safe RL training using each as a safety critic. The plot shows the results for both success rate (top row) and average return (bottom row) across different levels of action noise and external perturbations following the same experiment setting as in the main paper and as introduced in Sec.~\ref{app: domain_randomization}. Both methods show similar robustness in the action noise test, with only minor differences in performance as the noise level increases. However, when faced with external perturbations (especially on heightfield terrain), our method consistently outperforms the SafeFall predictor, demonstrating the superior generalizability and robustness of our receding-horizon safety critic under domain shifts. Our approach excels because it provides earlier and more accurate predictions of failure, which enables the safety layer to intervene more effectively and avoid catastrophic outcomes. In contrast, the SafeFall predictor tends to give more binary and delayed predictions, which can limit its effectiveness in handling complex, real-world disturbances. This highlights the advantage of using a state--action-based safety critic for environments with more dynamic and uncertain conditions.

\subsection{RL-PPO with Domain Randomization}
\label{app: domain_randomization}

We employ a set of domain randomizations to improve robustness and expose the safety critic to diverse dynamics during training. The randomizations are implemented as modular components and can be grouped into the following categories:
\begin{itemize}
    \item MuJoCo model randomization: randomizes physical parameters of the simulator model (e.g., masses, inertias, joint/actuator properties, and contact-related parameters) to induce dynamics variability.
    \item Observation noise: injects noise and/or bias into proprioceptive observations to emulate sensor corruption and partial observability.
    \item Action delay: applies stochastic control latency to account for delays between policy output and actuation.
    \item Joint dropout: randomly masks (drops) a subset of joint signals to simulate intermittent sensing/estimation failures.
    \item Initial-state randomization: perturbs the initial base pose/velocity and joint configuration at reset to diversify starting conditions.
    \item External perturbations: applies impulse-like disturbances directly to the base and joints. This is the primary shift we sweep at evaluation time and is detailed next.
\end{itemize}

Among the different domain randomization aspects, we elaborate in detail on how we do the evaluations with the controlled sim2sim shift on the external perturbations. The implementation details can be found in our codebase. 

\subsubsection*{External Perturbations (Evaluated Shift)}
The perturbation module injects impulse-like disturbances by directly modifying the simulator generalized coordinates and velocities $(q, \dot{q})$. In the evaluation setting, we apply the full-strength perturbation (i.e., the scaling coefficient is set to 1).

\paragraph{Base (Trunk) Velocity Perturbation}
We first define a magnitude bound for the 6D base velocity (3 linear + 3 angular) as
\begin{equation}
v_{\max} = \min\left( k_m \cdot M, v_{\mathrm{clip}} \right),
\end{equation}
where $M = \sum_b m_b$ is the total body mass, $k_m > 0$ is a mass-proportional scaling factor, and $v_{\mathrm{clip}} > 0$ is an absolute cap. We then sample a random base-velocity offset:
\begin{equation}
\Delta \dot{q}_{\mathrm{base}} \sim \mathcal{U}\left(-v_{\max}, v_{\max}\right)^6,
\end{equation}
and update the base velocity as:
\begin{equation}
\dot{q}_{\mathrm{base}}^{\text{new}} = \dot{q}_{\mathrm{base}} + \mathbf{1}\{\xi < p_{\mathrm{add}}\} \Delta \dot{q}_{\mathrm{base}},
\end{equation}
where $\xi \sim \mathcal{U}(0, 1)$ and $p_{\mathrm{add}} \in [0, 1]$ is the probability of applying the additive base-velocity perturbation. If the event does not trigger, the base velocity remains unchanged.

\paragraph{Joint Velocity and Position Perturbation}
For the actuated joints, we sample independent uniform perturbations for velocities and positions:
\begin{align}
\Delta \dot{q}_{\mathrm{j}} &\sim \mathcal{U}\left(-\bar{v}_{\mathrm{j}}, \bar{v}_{\mathrm{j}}\right)^{n_j}, \\
\Delta q_{\mathrm{j}} &\sim \mathcal{U}\left(-\bar{q}_{\mathrm{j}}, \bar{q}_{\mathrm{j}}\right)^{n_j},
\end{align}
where $n_j$ is the number of actuated joints, $\bar{v}_{\mathrm{j}} > 0$ bounds the per-joint velocity perturbation magnitude, and $\bar{q}_{\mathrm{j}} > 0$ bounds the per-joint position perturbation magnitude. The perturbed joint state is then updated as:
\begin{equation}
\dot{q}_{\mathrm{j}}^{\text{new}} = \dot{q}_{\mathrm{j}} + \Delta \dot{q}_{\mathrm{j}}, \qquad q_{\mathrm{j}}^{\text{new}} = q_{\mathrm{j}} + \Delta q_{\mathrm{j}}.
\end{equation}

In our evaluation sweeps, we double the magnitude of both base and joint perturbations and vary the perturbation sampling probability, which controls how frequently this procedure is applied during a rollout.

\subsubsection*{Action Noise}
\label{app:action_noise_test}
The action noise test evaluates the robustness of our safety critic against sub-optimal policy output. In this test, the magnitude of the action noise is sampled from a Gaussian distribution with a standard deviation equal to 50\% of the action range, ensuring that the noise perturbations are large enough to simulate significant errors while remaining within realistic operational bounds.

The key factor in this test is the variation in the \emph{sampling probability}, which controls how frequently the noise is applied to the action signals during the rollout. By performing this test, we are able to understand how the safety critic performs under noisy conditions, leading to sub-optimal policy outputs. This is a realistic scenario for many robotic applications, especially in dynamic and unpredictable environments. It also provides insight into the limits of the state-only versus state-action safety critic in handling increasing levels of disturbance, giving us a better understanding of their respective strengths and weaknesses in noisy, real-world settings.

 \begin{figure*}[ht]
\centering
\includegraphics[width=\textwidth]{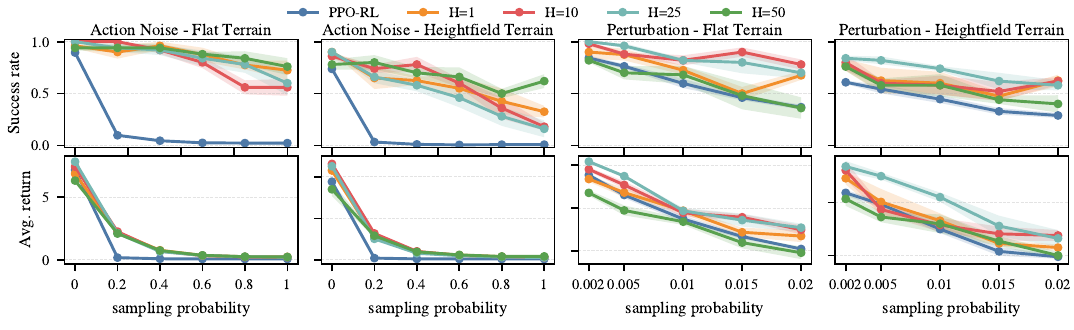}
\caption{Complete ablation on prediction horizon $H$. Success rate (top row) and average return (bottom row) under action noise and external perturbations on flat and heightfield terrain. The x-axis represents the action noise or disturbance sampling probability. Curves show mean performance over 5 seeds; shaded bands indicate standard error.}
\label{fig:app:exps_H_full}
\end{figure*}

\subsection{Ablation Study on Horizon Length}
\label{app:ablation_H}
This section reports the complete ablation results for the receding-horizon safety critic, varying the prediction horizon $H \in \{1, 10, 25, 50\}$. We evaluate under the same sim-to-sim shifts as in the main paper, including the action noise and impulsive external perturbations, on both flat and heightfield terrain. For each setting, we sweep the disturbance sampling probability and report success rate and average return (Fig.~\ref{fig:app:exps_H_full}). The PPO-RL baseline without shielding is included for reference.

Across all four suites, increasing disturbance frequency monotonically degrades performance, with the unshielded PPO-RL collapsing rapidly under action noise, especially on heightfields. Adding the predictive shield improves robustness for all $H$, but the horizon length governs a clear robustness-performance trade-off. Short horizons ($H{=}1$) behave more myopically and tend to lose success under stronger shifts, while very long horizons ($H{=}50$) can become overly conservative, leading to lower returns and, in some regimes, reduced success. A moderate horizon ($H{=}25$) is the most consistent choice in our experiments: it achieves the best or near-best success rates over a wide range of sampling probabilities, particularly under perturbations and on heightfields, while preserving higher returns than $H{=}50$. These trends mirror the discussion in the main paper and support selecting $H{=}25$ as a robust default across disturbance types and terrains.

\end{document}